# Recover Canonical-View Faces in the Wild with Deep Neural Networks

Zhenyao Zhu[1]     Ping Luo[1]     Xiaogang Wang[2]     Xiaoou Tang[1,3]

[1]Department of Information Engineering, The Chinese University of Hong Kong
[2]Department of Electronic Engineering, The Chinese University of Hong Kong
[3]Shenzhen Institutes of Advanced Technology, Chinese Academy of Sciences
`zz012@ie.cuhk.edu.hk`     `pluo.lhi@gmail.com`
`xgwang@ee.cuhk.edu.hk`    `xtang@ie.cuhk.edu.hk`

## Abstract

Face images in the wild undergo large intra-personal variations, such as poses, illuminations, occlusions, and low resolutions, causing great challenges to face-related applications. This paper addresses this challenge by proposing a new deep learning framework that can recover the canonical view of face images. It dramatically reduces the intra-person variances, while maintaining the inter-person discriminativeness. Unlike the existing face reconstruction methods that were either evaluated in controlled 2D environment or employed 3D information, our approach directly learns the transformation from the face images with a complex set of variations to their canonical views. At the training stage, to avoid the costly process of labeling canonical-view images from the training set by hand, we have devised a new measurement to automatically select or synthesize a canonical-view image for each identity.

As an application, this face recovery approach is used for face verification. Facial features are learned from the recovered canonical-view face images by using a facial component-based convolutional neural network. Our approach achieves the state-of-the-art performance on the LFW dataset.

## 1 Introduction

Dealing with variations of face images is the key challenge in many face-related applications. For example, in face recognition, most research efforts have focused on how to distinguish intra-personal variations of poses, lightings, expressions, occlusions, ages, makeups, and resolutions from inter-personal variation which distinguishes face identities. The aim of face hallucination is to reconstruct high-resolution face images from low-resolution ones [26], or to remove glasses from face images [39]. For face synthesis, people produce images under different ages [32], poses [8], and illuminations [36]. There are also research works [38] on matching face photos with sketches of different styles and synthesizing sketches from photos. Recently, a 3D viewing system [18] was proposed to reconstruct 3D face models from real-world images.

To deal with face variations, the existing methods can be roughly divided into two categories: robust feature extraction and face normalization. In the first category, global features such as Eigen faces [33], Fisher faces [6], and their extensions [37] can cover global variations due to small pose and simple illumination changes, but do not work well under large poses and complex illumination conditions. They are not robust to local distortions, such as expressions and occlusions either. The high dimensional concatenations of the local descriptors, such as Haar [35], Gabor [15], and LBP [1], have demonstrated their robustness to local distortions and achieved significant improvement in face recognition [11]. In addition to the above hand-crafted descriptors, other existing researches studies have also tried to integrate multiple features or directly learn features from raw pixels, such as using random-projection trees [10], local quantized patterns [34], and deep learning [19, 24, 21, 13, 30, 31]. For example, Sun et al. [31] learned face representation with a deep model through face identification, which is a challenging multi-class prediction task. The common weakness of the feature extraction approaches is that they are all sensitive to large intra-person variations.

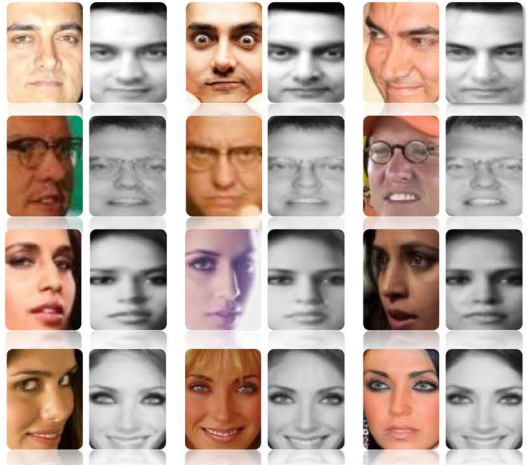

Figure 1: The proposed method can recover the images of canonical view and illumination from images with large variations. For example, in each row, we show the images and the reconstructed images of the same identity. The reconstructed images dramatically reduce the intra-person variances, while maintaining the inter-person discriminativeness.

In the second category, approaches tend to recover an image in the canonical view (with frontal pose and neutral lighting) from a face image under a large pose and a different lighting, so that it can be used as a good normalization. There are 3D- and 2D-based methods. The 3D-based methods aim to recover the frontal pose by 3D geometrical transformations [8, 4], which first aligns a 2D face image to a 3D face model and then rotates it to render the frontal view. The existing 2D-based methods [3, 2] inferred the frontal pose with graphical models, such as Markov Random Fields (MRF), where the correspondences between nodes in the MRF are learned from images in different poses. However, capturing 3D data adds additional cost and resources, and MRF-based face synthesis depends heavily on good alignment, while the results are often not smooth on real-world images. The recent work [40] directly learned transformation between face images in arbitrary views and frontal views and obtained good results in the MultiPIE dataset [17].

In this paper, we aim to recover the canonical view from a 2D face image taken under an arbitrary pose and lighting condition in the wild. It is a big challenge to learn such a complex set of pose and lighting transforms in uncontrolled environment, and the learned transformation function must be highly multi-modal. We will show that a carefully designed deep learning framework can overcome this challenge, benefiting from its great learning capacity. Some examples of recovered face images with our approach are shown in Figure 1.

Our framework contains two steps: (1) canonical-view image selection, and (2) face recovery. First, in order to learn the transformation between face images and their canonical views, we must select a representative image for each identity, which is taken in the frontal view, under neutral lighting condition and with high resolution. To avoid selecting them by hand, we first develop a new measurement, which measures the face images' symmetry and sharpness. We then learn the transformation with a carefully designed deep network, which can be considered as a regression from images in arbitrary views to the canonical-view.

A sample application of this framework is face verification. A facial component-based convolutional neural network is developed to learn hierarchical feature representations from the recovered canonical-view images. These features are robust for face verification, since the recovered images already remove large face variations. It also has potential applications to other problems, such as face hallucination, face sketch synthesis and recognition, and face attribute estimation.

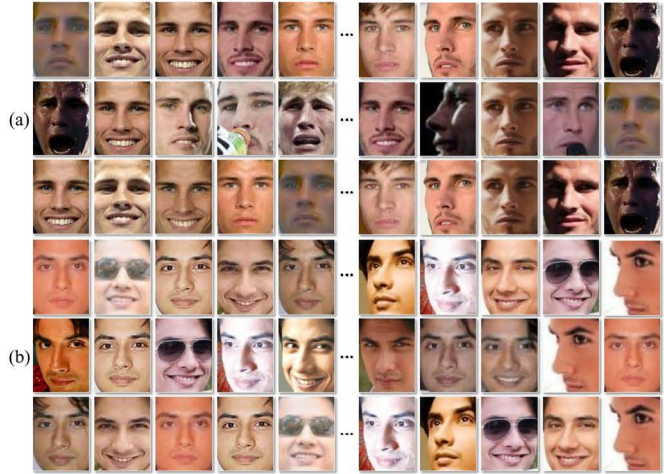

Figure 2: The images of two identities are ranked according to three different criteria in (a) and (b): face symmetry (the first row), matrix rank (the second row), and symmetry combined with matrix rank (the third row). In each row, the first five images and the last five images are visualized.

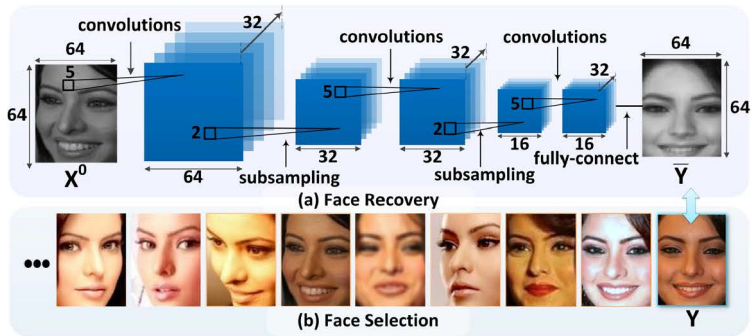

Figure 3: Pipeline of canonical view face selection (b) and face recovery (a).

In summary, this work has the following key **contributions**. Firstly, to the best of our knowledge, this is the first work that can recover canonical-view face images using only 2D information from face images in the wild. This method shows state-of-the-art performance on face verification in the wild. Secondly, the reconstructed images are of high-quality.

## 2 Canonical View Face Recovery

### 2.1 A New Measurement for Canonical View Face Images

Although various facial measurements [29] have been studied in the literature, they have mainly focused on the image qualities, such as noise ratio and resolutions, and seldom considered how to determine whether a face image is taken in frontal view. We have devised a facial measurement for frontal view face images by combining the rank and symmetry of matrix. For example, as shown in Figure 2, we collect the images of a subject and visualize them according to the following three criteria: (1) difference between the left half face and the right half face in ascending order (face symmetry), (2) rank of the image in descending order, and (3) the combination of (1) and (2). In the first row of Figure 2, we observe that measuring symmetry as in (1) is effective for frontal view images, but it prefers the images in low resolutions. Although the second row shows that larger rank indicates sharper images, the images sometimes do not have frontal views. The combination of (1) and (2) achieves the best result as shown in the third row in Figure 2.

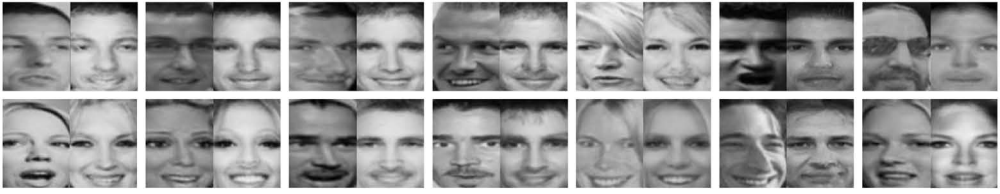

Figure 4: Examples of face reconstruction on the LFW dataset. For each pair, the left one is an original image of the LFW dataset and the right one is the recovered image.

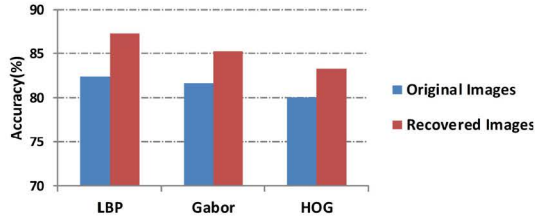

Figure 5: Comparisons of face verification performance of different features on the LFW dataset.

We formulate this measurement as shown below. Let a matrix $\mathbf{Y}_i \in \mathbb{R}^{64 \times 64}$ denote a face image of the $i$-th identity and $\mathbb{D}_i$ be the set of images of identity $i$, $\mathbf{Y}_i \in \mathbb{D}_i$. The frontal view measurement can be written as,

$$M(\mathbf{Y}_i) = \| \mathbf{Y}_i \mathbf{P} - \mathbf{Y}_i \mathbf{Q} \|_F^2 - \lambda \| \mathbf{Y}_i \|_*, \tag{1}$$

where $\lambda$ is a constant coefficient, $\| \cdot \|_F$ is the Frobenius norm, and $\| \cdot \|_*$ denotes the nuclear norm, which is the sum of the singular values of a matrix. $\mathbf{P}, \mathbf{Q} \in \mathbb{R}^{64 \times 64}$ are two constant matrixes with $\mathbf{P} = diag([\mathbf{1}_{32}, \mathbf{0}_{32}])$ and $\mathbf{Q} = diag([\mathbf{0}_{32}, \mathbf{1}_{32}])$, where $diag(\cdot)$ indicates the diagonal matrix. The first term in Equation (1) measures the face's symmetry, which is the difference between the left half and the right half of the face, and the second term measures the rank of the face. Smaller value of Equation (1) indicates the face is more likely to be in frontal view.

We can select a frontal face image as a representative for each identity and then learn a mapping, which transforms the face image in arbitrary view to the frontal view. This selection can be achieved in several ways. In this report, we simply choose the image with the minimum measurement as the frontal face for each identity. However, using a linear combination to calculate the frontal face image is also possible. We will report results in the future.

### 2.2 Face Recovery

After face selection, we adopt a deep network to recover the frontal view image by minimizing the loss error

$$E(\{\mathbf{X}^0_{ik}\}; \mathbf{W}) = \sum_i \sum_k \| \mathbf{Y}_i - f(\mathbf{X}^0_{ik}; \mathbf{W}) \|_F^2, \tag{2}$$

where $i$ is the index of identity and $k$ indicates the $k$-th sample of identity $i$. $\mathbf{X}^0$ and $\mathbf{Y}$ denote the training image and the target image (the selected frontal face), respectively. $\mathbf{W}$ is a set of parameters of the deep network.

As shown in Figure 3, the deep network contains three convolution layers. The first two are followed by the max pooling layers and the last one is followed by a fully-connect layer. Different from the conventional CNN, whose filters share weights, our filters are localized and do not share weights because we assume different face regions should employ different features. The input $\mathbf{X}^0$, the output $\overline{\mathbf{Y}}$ (predicted image), and the target $\mathbf{Y}$ are in the size of $64 \times 64$. All of them are transformed to gray-scale and their illuminations are corrected as in [36]. At each convolutional layer, we obtain 32 output channels by learning non-shared filters, each of which is in the size of $5 \times 5$. The cell size

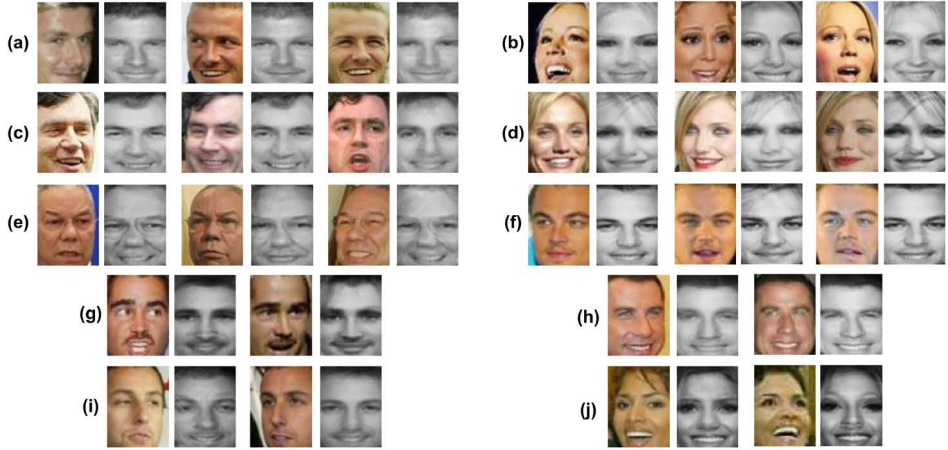

Figure 6: Canonical view face reconstructions of several identities.

of the sub-sampling layer is $2 \times 2$. The $l$-th convolutional layer can by formulated as

$$\mathbf{X}^{l+1}_{q,uv} = \sigma(\sum_{p=1}^{I} \mathbf{W}^l_{pq,uv} \circ (\mathbf{X}^l_p)_{uv} + \mathbf{b}^l_q),  \qquad(3)$$

where $\mathbf{W}^l_{pq,uv}$ and $(\mathbf{X}^l_p)_{uv}$ denote the filter and the image patch at the image location $(u, v)$, respectively. $p, q$ are the indexes of input and output channels. For instance, in the first convolutional layer, $p = 1, q = 1...32$. Thus, $\mathbf{X}^{l+1}_{q,uv}$ indicates the $q$-th channel output at the location $(u, v)$; that is the input to the $l+1$-th layer. $\sigma(x) = \max(0, x)$ is the rectified linear function and $\circ$ indicates the element-wise product. The bias vectors are denoted as $\mathbf{b}$. At the fully-connect layer, we recover the image $\overline{\mathbf{Y}}$ by

$$\overline{\mathbf{Y}} = \mathbf{W}^L \mathbf{X}^L + \mathbf{b}^L. \qquad(4)$$

Equation (2) is non-linear because of the activation functions in the deep network. We solve it by the stochastic gradient descent (SGD) with back-propagation as in [22]. As shown in Figure 3, at the $l$-th convolutional layer, the gradient of the filter at position $u, v$ is computed by $\frac{\partial E}{\partial \mathbf{W}^l_{uv}} = (\mathbf{e}^l)_{uv}(\mathbf{X}^{l-1})_{uv}$, where $E$ is the loss error defined in Equation (2) and $\mathbf{e}$ is the back-propagation error. $\mathbf{e}$ is obtained in a recursive manner as $\mathbf{e}^l = f^{l'} \circ (\mathbf{e}^{l+1} \otimes \mathbf{1})$, where $\otimes$ is the Kronecker product that up-samples $\mathbf{e}^{l+1}$ to the same size as $\mathbf{e}^l$, and $f^{l'}$ is the derivative of the activation function at the $l$-th layer. At the $l$-th fully-connect layer, the gradient of the weight matrix is calculated by $\frac{\partial E}{\partial \mathbf{W}^l} = \mathbf{X}^{l-1}\mathbf{e}^{l^T}$, which is the outer product of the back-propagation error and the input of this layer. $\mathbf{e}$ is also derived in a recursive way as $\mathbf{e}^l = f^{l'} \circ (\mathbf{W}^{l+1^T}\mathbf{e}^{l+1})$. For instance, if layer $l$ is activated using sigmoid function, then $\mathbf{e}^l = \mathbf{X}^l \circ (\mathbf{1} - \mathbf{X}^l) \circ (\mathbf{W}^{l+1^T}\mathbf{e}^{l+1})$. Furthermore, dropout learning [20] is adopted at each layer to avoid over-fitting.

## 2.3 Effectiveness of Face Recovery

Several examples of the recovered canonical view images are shown in Figure 4. In order to demonstrate the quality of the recovered image, we compare the performance of the existing feature extraction methods, including LBP [1], HOG [14], and Gabor [15], when they are extracted from the reconstructed image and the original image. We adopt the testing data of LFW dataset. For each of the above features, we extract it from the face image in a regular grid of size $8 \times 8$ and then apply PCA and LDA. The performance of face verification are reported in Figure 5, where shows that the existing feature extraction methods can be improved when they are applied on the recovered image, which is a good normalization to account for different face variations. More examples of face recovery for one identity can be found in Figure 6.

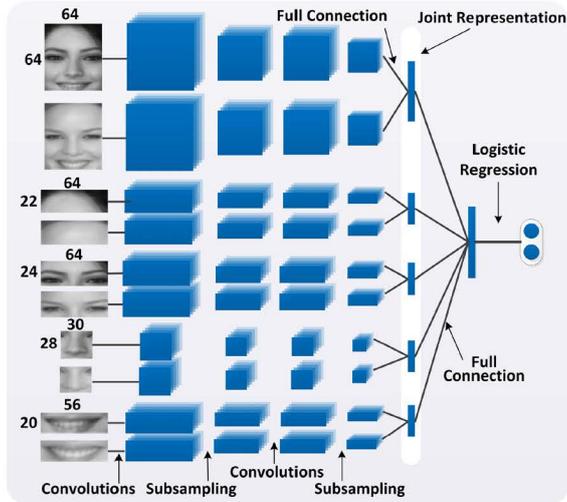

Figure 7: Architecture of the facial component-based network. The network contains five CNNs, each of which takes a pair of whole faces or facial components as input. The sizes of the whole face, forehead, eye, nose, and mouth are 64 × 64, 22 × 64, 24 × 64, 28 × 30, and 20 × 56, respectively. First, each CNN learns the joint representation of the pairs of input. A logistic regression layer then concatenates all the joint representations as features to predict whether the two face images belong to the same identity.

## 3 Facial Component Deep Network for Face Verification

The canonical view images can be used as input to a facial component deep network (FCN), which learns relational features from two images for face verification, as shown in Fig.7. Similar architecture has been adopted by [30], where the original images are used as the input and a large number of networks have to be trained. Unlike [30], the FCN is applied on the canonical images that reduce the face variations. Therefore, five networks concatenation is enough to achieve good result as discussed below. Learning FCN contains three steps, including facial component-based patch recovering and cropping, feature learning, and feature reduction.

In the first step, for each pair of training images, we recover their canonical view images and then extract 5 landmarks. Image patches of different facial components are cropped based on the above landmarks. Specifically, we extracted patches from forehead, eyes, nose, and mouth.

In the second step, each patch pair is utilized to train a deep network. Then, multiple networks are concatenated together by a fully-connected layer to learn the feature representation. Each network comprises two convolutional layers and two sub-sampling layers. Figure 7 specifies the architecture of concatenation of multiple networks, where the parameters are optimized using stochastic gradient descent with back-propagation. In particular, as in Section 2.2, we pass the back-propagation error backwards and then update the weights or filters in each layer. We adopt the entropy error instead of the loss error because we need to predict the labels $\mathbf{y}$

$$Err = \mathbf{y} \log \overline{\mathbf{y}} + (1 - \mathbf{y}) \log(1 - \overline{\mathbf{y}}), \qquad (5)$$

where $\overline{\mathbf{y}}$ are the predicted labels, and $\mathbf{y}, \overline{\mathbf{y}} \in \{0, 1\}^K$, with $\mathbf{y}_k = 1$ indicating that the input images belong to the $k$-th identity.

In the third step, we employ PCA and ensemble of Support Vector Machines (SVM) for face verification.

## 4 Experiments

We evaluate our approach on the LFW dataset, which is collected from internet and contains 5749 people with 13, 233 face images in total, which vary in terms of their poses, illuminations, resolutions, makeups, and occlusions. The average number of images for each identity is 2.3±9.01, where 5438 people have less than 5 images and only 143 people have more than 10 images. Due to the imbalance of LFW, it is not suitable to train the face recovery network because of the following reasons: (1) training examples are not enough for most of the identities, (2) they may not have frontal view images, and (3) the size of the dataset is not enough for a deep learning-based method. PubFig [23] and WDRef [12] are two larger datasets than LFW. However, PubFig only has 200 people, which means the identity variation is insufficient, while WDRef is not publicly available. We train our models on the CelebFaces [30], which contains $87,628$ face images of $5436$ identities. The average number of images for each identity is $15.9\pm8.0$, which shows that it is more balanced than LFW.

| Methods | Accuracy(%) |
|---|---|
| Associate-Predict [28] | 90.57 |
| Joint Bayesian† [12] | 92.42 |
| Convnet-RBM [30] | 92.52 |
| Tom-vs-Pete [7] | 93.10 |
| Tom-vs-Pete+Attribute [7] | 93.30 |
| High-dim LBP† [11] | 95.17 |
| TL Joint Bayesian [9] | 96.33 |
| FR+FCN† (whole face+components) | 96.45 |
| PLDA [25] | 90.07 |
| Joint Bayesian [12] | 90.90 |
| Fisher Vector Faces [5] | 93.03 |
| High-dim LBP [11] | 93.18 |
| FR+FCN (whole face) | 93.65 |
| FR+FCN (whole face+components) | 94.38 |
| Face++ [16] | 97.27 |

Table 1: Method Comparisons.

We compare our results with the existing best-performing approaches suggested by the LFW benchmark[1]. There are two experimental settings. First, the upper part of Table 1 shows the results employing outside training data other than LFW under the restricted protocol. Most of the best-performing methods such as [9, 11, 7] belong to the second setting. Second, the methods in the lower part are trained on LFW under the unrestricted protocol, using only the training data in LFW.

Our methods achieve the state-of-the-art performance in both the above settings. For instance, in the first setting, we train the FR+FCN† (whole face+components) on the outside data of two hundred thousand image pairs generated from the PubFig [23] and CelebFaces. The FR+FCN (whole face+components) achieves the accuracy of 96.45 percent, which performs slightly better than the best results [9] and improves 4 percent compared to [30]. This is because the canonical view images can reduce large face variations. In the second setting, we achieve the second best result. The best-performing method is a commercial system [16], where the number of facial landmark alignment and the size of training data are not clear. Our method employs the recovered the canonical view images to reduce the face variations. In this case, five facial key points alignment is enough to achieve good result. Figure 8 and 9 plots the ROC curves of the above methods. For more details please refer to the project page at http://mmlab.ie.cuhk.edu.hk.

## 5 Conclusions and Discussions

In this paper, we have proposed a new deep learning framework that can recover the canonical view face images from images in arbitrary wild conditions. With this framework, given the face images of any new identity, the canonical view of these images can be efficiently recovered. This approach has many potential applications, such as face hallucination, face sketch synthesis and recognition, and face attribute estimation.

---
[1] http://vis-www.cs.umass.edu/lfw/results.html

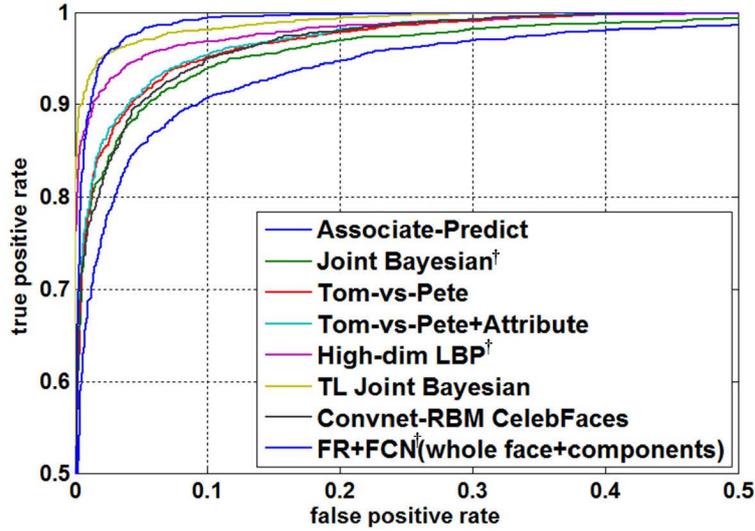

Figure 8: ROC curve under the LFW restricted protocol.

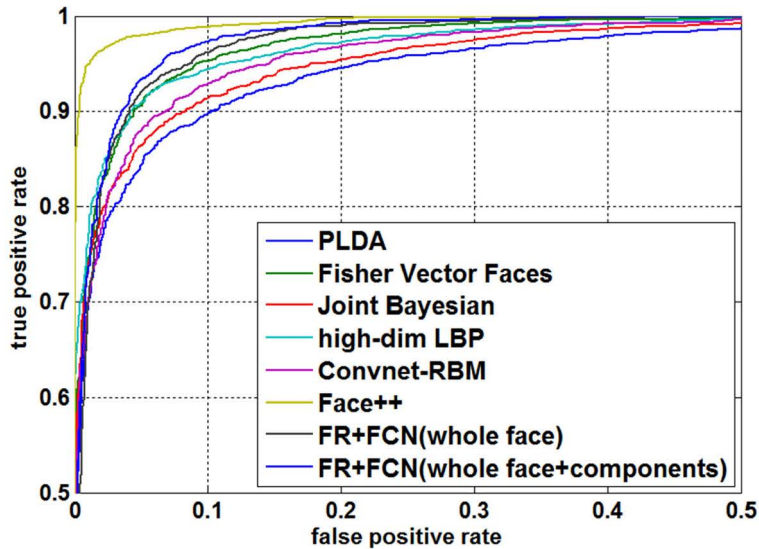

Figure 9: ROC curve under the LFW unrestricted protocal

We apply our face recovery framework to the task of face verification and outperform the state-of-the-art approaches. We also show that the existing face recognition methods can be improved when they adopt our face recovery as normalization and pre-processing.

A recent work [27] reported 98.5 percent accuracy with Gaussian Processes and combined multiple training sets. This could be due to fact that the nonparametric Bayesian kernel method can adapt model complexity to data distribution. This could be another interesting direction to be explored in the future.

## 6 Acknowledgement

This work is partially supported by the General Research Fund sponsored by the Research Grants Council of Hong Kong (Project No.CUHK 416510 and 416312) and Guangdong Innovative Research Team Program (No.201001D0104648280).